# Evaluation The Efficiency of Artificial Bee Colony And The Firefly Algorithm In Solving The Continuous Optimization Problem

Seyyed Reza Khaze[1], Isa maleki[2], Sohrab Hojjatkhah[3] and Ali Bagherinia[4]

[1]Department of Computer Engineering, Dehdasht Branch, Islamic Azad University, Iran,
`khaze@iaudehdasht.ac.ir, khaze.reza@gmail.com`

[2]Department of Computer Engineering, Dehdasht Branch, Islamic Azad University, Iran,
`maleki@iaudehdasht.ac.ir, maleki.misa@gmail.com`

[3]Department of Computer Engineering, Dehdasht Branch, Islamic Azad University, Iran,
`hojjatkhah@gmail.com`

[4]Department of Computer Engineering, Dehdasht Branch, Islamic Azad University, Iran,
`ali.bagherinia@gmail.com`

## Abstract

*Now the Meta-Heuristic algorithms have been used vastly in solving the problem of continuous optimization. In this paper the Artificial Bee Colony (ABC) algorithm and the Firefly Algorithm (FA) are valuated. And for presenting the efficiency of the algorithms and also for more analysis of them, the continuous optimization problems which are of the type of the problems of vast limit of answer and the close optimized points are tested. So, in this paper the efficiency of the ABC algorithm and FA are presented for solving the continuous optimization problems and also the said algorithms are studied from the accuracy in reaching the optimized solution and the resulting time and the reliability of the optimized answer points of view.*

## 1. Introduction

Now the use of the Meta-Heuristic algorithms in accessing the optimized solution in the continuous optimization problems has progressed a lot. According to the increase of the complexity of the continuous optimization problems and the inability of the mathematical methods for the optimized solution, the Meta Heuristic algorithms are the suitable solution for the continuous optimization problems. The mathematical methods are used in many scientific and engineering problems and cover a vast area of the different problems but despite the accurate efficiency, the mathematical methods still face many problems solving the optimization problems. The late researches and the struggles of the researchers have led to innovation of the algorithms which have been inspired by the natural phenomenon, the ones which study the completion and the behavior of the creatures of the nature and finally they have led to the Met-Heuristic algorithms. The Meta-Heuristic algorithms have been efficient in solving the combined optimization problems in finding the optimized solution [1, 2, 3 and 4].

Many Meta-Heuristic algorithms have been innovated inspiring the nature of which the Particle Swarm Optimization (PSO) [5], Artificial Bee Colony (ABC) [6], Firefly Algorithm (FA) [7], Bee Colony Optimization (BCO) [8] and Ants Colony Optimization (ACO) [9] could be pointed out. The ABC algorithm [6, 10] is a Meta-Heuristic algorithm which is inspired by the mining behavior of the bee colony for solving the continuous optimization problems of large space. ABC algorithm starts working by creating the primary population of the random vectors. It works in a way that in any repetition of the algorithm, the artificial bees randomly search the answers which have been resulted in the previous repetition to find new answers. It is clear that the new answers would not be necessarily better than the answers found in previous repetition. When any of the artificial bees find a new answer, they will go back to the hive and will make decision for the next move path in the next repetition. So, the optimization rate of the answers is calculated by the bees and then the answer which is more befitting, will be selected as the search path in next repetition. So, the area around the more optimized answers will be searched by more bees in the next repetition. The search process continues until the needed conditions for ending the execution of the program would be met.

The FA [7] is one of the newest Meta-Heuristic algorithms based on the swarm intelligence which is used in solving the continuous optimization problems. First some artificial firefly are randomly distributed in the problem space in FA, and then any firefly emit light, the intensity of which is in conformity to the optimization rate of the point the firefly stands on. Then the light intensity of any firefly is compared to the light intensity of the fireflies and the low light firefly goes toward the intense lighted firefly. Also the most intense firefly moves around the problem for finding the global optimized answer randomly. So, in FA the fireflies get in relationship with each other via the light. The combination of these operations leads to the movement of the all fireflies toward the more optimized points. In this paper, we study the ABC algorithm and FA and to show the efficiency of these algorithms, will solve some continuous optimization functions.

The structure of the paper is as follows: in the section 2, we have introduce the related works; in the section 3, the Meta-Heuristic algorithm is introduced; in the section 4, the analysis of ABC and FA are for solving the continuous optimization problems has been studied; in the section 5, we have evaluated the results of ABC and FA are; in the section 6, ABC and FA are discussed and at finally in section 7, conclusion and future works is presented.

## 2. RELATED WORKS

X. Lia [11] has used the Particle Swarm Optimization (PSO) Algorithm and Genetic Algorithm (GA) to solve the continuous optimization problems. He has studied the PSO and GA algorithms to test and evaluate the efficiency factor on 36 functions. To clearly show the efficiency of the PSO and FA algorithms, he tested the functions in 30 dimensional spaces. The results of the tests show that these algorithms have worked well on the 30 functions and found the optimized answers. The researchers [12] have used the Dynamic PSO and Simple PSO to solve the continuous mathematical functions. They have studied the functions in the 2 and 10 dimensional spaces to show the efficiency of both algorithms. They have studied the parameters of the Dynamic PSO and Simple PSO to analyze the mathematical functions and they have resulted that Dynamic PSO is more efficient for solving the continuous problems and creates the answers close to the optimized one. Reference [13] has used the ABC algorithm and the Differential Evolution (DE) for solving the optimization problems in large scale. In this reference, a combined method named DEM-ABC has been suggested. In the combined method for global convergence of the ABC Algorithm, the DE Mutation Strategy is used. The combined algorithm is tested on some functions in large scales. The results show that the combined algorithms more efficient than the ABC Algorithm.

Researchers [14] have used ABC algorithm and Hybrid Artificial Bee Colony (HABC) Algorithms for solving the continuous optimization problems. The DE algorithm is used for optimizing the answers of the continuous optimization functions in HABC. They have tested the ABC and HABC algorithms for evaluation and efficiency factors on 6 functions in the paper. To show the efficiency of the ABC and HABC algorithms, they have studied the functions in 30 and 60 dimensional spaces. The results show that HABC is more efficient than ABC algorithm.

Researchers [15] have used FA to solve the non-linear continuous functions. The results of the tests show that FA is fast enough in convergence toward the optimized solution and finds the optimized answer in a very short time. Reference [16] has used the ABC algorithm for solving the continuous optimization functions. Improving the ABC algorithm, it is tried to go through Exploitation and Local Search operations for reaching the optimized the answer in this reference. The results show that the improve ABC algorithm has reached more optimized answer than the ABC algorithm.

Researchers [17] have used the Bacterial Foraging Optimization Algorithm (BFOA) to study and analyze the continuous functions. They have showed the efficiency of the BFOA by solving some continuous functions. The goal of these functions is to find the optimized answers in multi-dimensional spaces. They have also studied the effects of the number of the bacteria on solving the functions and have cited that accurate identification of the parameters makes BFOA be very effective in optimizing the problems. Reference [18] has used the ABC algorithm to solve the continuous optimization functions. To make better global search in this reference, some changes has took place on exploration and exploitation operations using DE algorithm. The goal was to make better global search and make ABC algorithm reach the answer as fast as possible. The results of the tests show that the algorithm is able to find the optimized solution. The researchers [19] have used FA and PSO algorithms to solve the continuous optimization problems. They have cited that the PSO algorithm is very efficient in exploration and exploitation operations of the continuous optimizations problems and also the FA is not very complex and is able to find the optimized answer in a very short time.

## 3. META-HEURISTIC ALGORITHMS

The Meta-Heuristic algorithms are the tools for finding the answers or the answers close to the optimized ones [20]. These algorithms utilize the two concepts of searching and cooperation search the searching space of an optimization problem. So, the more powerful is an algorithm in controlling these two parameters, the more able is the algorithm in finding the answers close to the optimized one for the problem.

### 3.1 ABC

The ABC algorithm is innovated in 2005 by Karaboga inspiring the social life of the bees to solve the optimization problems [6]. This algorithm is a simulation of the food search of the group of the bees. The bees can be distributed in far and different distances to utilize the food resources [21]. In ABC algorithm, the bees are classified in three groups: 1. Employed bees, 2. Onlooker bees, 3. Scout bees.

The food search process starts by the employed bees. Each employed bee dances in a specific way when finds food resource and the onlooker bees look at the dance of the employed bees to understand the food resource location and the scout bees randomly look for the food in the around environment.

In ABC algorithm the primary value stages of the employed, onlooker and scout bees are as follows:

**Employed bees:** In this stage the artificial bees searching around the food resource at $x_i$ point will search for the better food resource at new location of $v_i$ [22]. Identification of the new food resource takes place by the equation (1) [23].

$$v_{ij} = x_{ij} + \phi_{ij}(x_{ij} - x_{kj}) \qquad j = 1,2,...,n; \quad k = 1,2,...,SN$$

(1)

In equation (1), $v_i = [v_{i1}, v_{i2}, ... v_{in}]$ is the new location vector of the bees, $x_i = [x_{i1}, x_{i2}, ..., x_{in}]$ is the location vector of the $i^{th}$ bee, k ($k \neq j$) is a correct random number in [1, SN] and the SN is the number of the artificial bees. $\Phi_{ij}$ is a random number uniformly distributed in [-1, 1]. The random $x_i$ number selection from the problem limit is done by the equation (2) [23].

$$x_{ij} = L_j + rand(0,1) \times (U_j - L_j)$$

(2)

In equation (2), $U_j$ and $L_j$ are the top limit and the down limit of the $x_i$ variable respectively and the rand() is the random numbers function in (0, 1). When the new location of the food resource is identified, the optimization of it must be calculated. So, the befitting rate of the $x_i$ vector is identified according to the equation (3) [23].

$$fit_i = \begin{cases} \dfrac{1}{1+f_i} & f_i \geq 0 \\ 1+abs(f_i) & f_i < 0 \end{cases}$$

(3)

**Onlooker bees**: In this stage any of the onlooker bees decide to search around the found food resource by the specified possibility [22]. The onlooker bees make their selection according to the possible values of the employed bees. So, the possibility of selection of the food resource by the onlooker bees is calculated using the equation (4) [23].

$$p_i = \dfrac{fit_i}{\sum\limits_{j=1}^{SN} fit_i}$$

(4)

**Scout bees**: In ABC algorithm, if the known number of the repetitions would not lead to the optimized answer, some of the bees leave their solution and become scout bees to randomly search the limits of the problem for increasing the search process efficiency [22]. Execution of the scout bees' stage can increase the possibility of finding the global optimized answer.

### 3.2 FA

FA is of the algorithms based on the population which is introduced in 2008 by Yang [7]. First a number of artificial bees are randomly produced in problem space in FA. Then a light intense is related to any of the fireflies using the value found for the goal function at that point. The light intense valuing of any of the artificial fireflies is in a way that increasing the optimization rate of the point a firefly stands on leads to increase of the light intense of it. The low light fireflies are attracted toward the high light fireflies and this continues till all fireflies are gathered in one point which is probably the global optimized point. Updating the law of the movement of the low light bees toward the high light ones takes place using equation (5) [24].

$$x_i \leftarrow x_i + \beta_0 e^{-\gamma r_{ij}^2}(x_j - x_i) + \alpha(rand - \frac{1}{2})$$

(5)

In equation (5), the values of $\alpha$, $\beta_0$ and $\gamma$ are considered constant. $\alpha$, $\beta_0$ are selected from [0, 1] and $\gamma$ is selected from [0, $\infty$). Also, $r_{ij}$ is the Euclid distance of the two fireflies which is identified as equation (6) [24].

$$r_{ij} = \|x_i - x_j\| = \sqrt{\sum_{k=1}^{n}(x_{i,k} - x_{j,k})^2}$$

(6)

The absorption coefficient between two fireflies takes place using equation (7).

$$\beta = \beta_0 e^{-\gamma r_{ij}}$$

(7)

In equation (7), $\beta 0$ shows the maximum of absorption and is identified in [0, 1]. The $\gamma$ parameter is the absorption coefficient and is identified in [0, $\infty$). The r parameter identifies the distance between the fireflies and the value of it is calculated by equation (6). If $\beta_0$=0, any of the fireflies searches the problem space without any contribution of the other fireflies and the search takes place randomly. Also, if $\gamma=\infty$, this leads to the random search in problem space.

## 4. EVALUATION THE EFFICIENCY OF ABC AND FA FOR SOLVING THE CONTINUOUS OPTIMIZATION FUNCTIONS

Searching the continuous optimization functions using the Meta-Heuristic algorithms leads to the most optimized solution among the possible solutions. The Meta-Heuristic algorithms in solving the continuous optimization problems are very efficient in getting close to the optimized answer. These algorithms study the continuous optimization problems answers using the testing and the better searching methods according to the problem structure and the complexity type of it and produce accessing the optimized answers.

### 4.1 ABC in Solving the Continuous Optimization Functions

The goal of the optimization is to find the optimized solutions according to the limits and the needs of the problems. It is possible for an optimization problem to have different solutions and for selecting the optimized answer the goal function must be used. The flowchart of evaluation of the goal function of the ABC algorithm for continuous optimization problems is showed in figure (1).

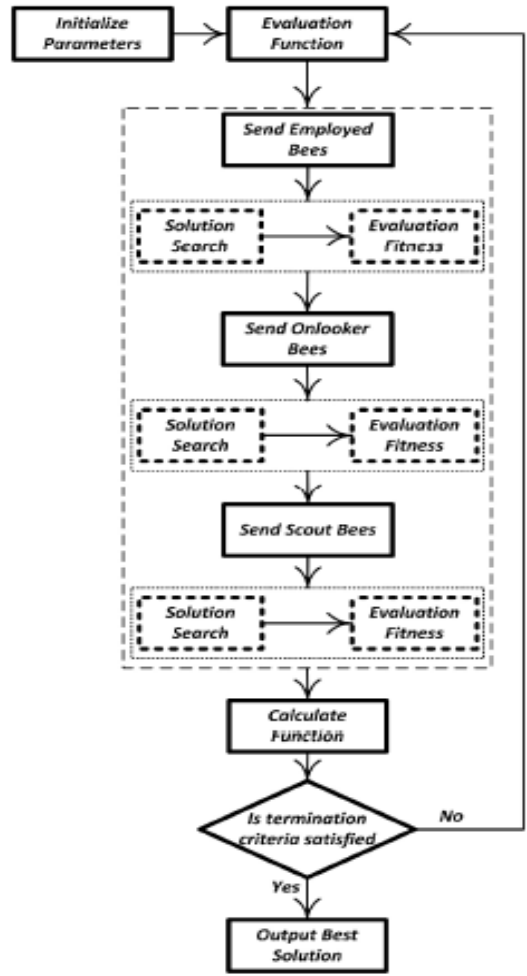

Figure 1: The Flowchart of ABC Algorithm for Solving the Continuous Optimization Functions

Figure (2) shows the quasi code of ABC algorithm for solving the continuous optimization Functions.

```
1. Initialize Parameters
2. Do
3. Evalution Function
4. Evalution Fitness
5. Employed Bee
For each employed bee do
Select a neighbor employed bee randomly
Update position
Calculate the fitness
Select the better solution
6. Onlooker Bee
For each onlooker bee do
Select a neighbor employed bee randomly
Update position
Calculate the fitness
Select the better solution
7. Scout Bee
For each scout bee do
Solution search
Update position
Calculate the fitness
Select the better solution
8. While (a stop criteria maximum iteration)
```

Figure 2: The Quasi Code of ABC Algorithm for Solving the Continuous Optimization Functions.

## 4.2 FA in Solving the Continuous Optimization Functions

Evaluation flowchart of the FA for solving the continuous optimization functions is shown in figure (3).

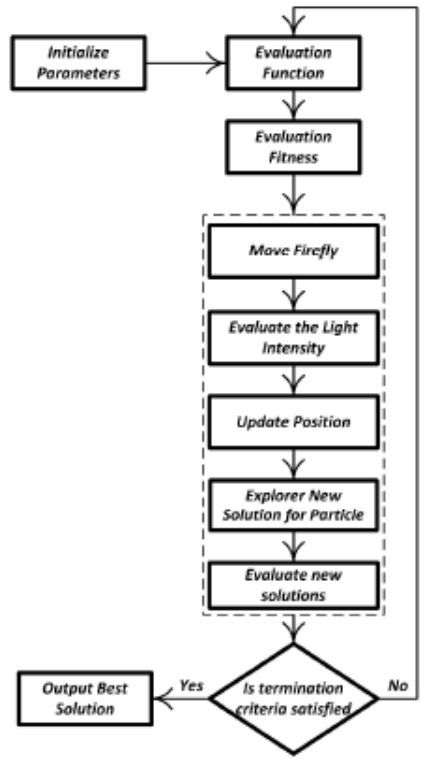

Figure 3: Flowchart of the FA for Solving the Continuous Optimization Functions

Figure (4) shows the FA quasi code for solving continuous optimization functions

```
1. Initialize Parameters
2. Do
3. Evaluation Function
4. Evaluation Fitness
5. The Algorithm of Firefly Algorithm
        for i=1 to n do
         for j= 1 to i do
            Move firefly i towards j
            Move firefly i randomly
            Evaluate new solutions
        End for j
        End for i
      Rank the fireflies and find the current best
6. While (a stop criteria maximum iteration)
```

Figure 4: FA Quasi Code for Solving Continuous Optimization Functions

## 5. EVALUATION AND RESULTS

For evaluation and efficiency of the ABC and FA are, the Rastrigin two and three dimensional function is studied [25]. This function holds many maximum and minimum points which have made it be used as a test function for evaluation of the Meta-Heuristic Algorithms. So, for comparison of the evaluation and efficiency of these algorithms, the Rastrigin function based on optimized solution factor is used.

- Two dimensional Rastrigin function

$$f_1(x) = 20 + \sum_{i=1}^{2} [x_i^2 - 10\cos(2\pi x_i)]$$

- Three dimensional Rastrigin function

$$f_2(x) = 30 + \sum_{i=1}^{3} [x_i^2 - 10\cos(2\pi x_i)]$$

The Meta-Heuristic algorithms are very delicate for their parameters and the settlement of the parameters can affect their operation. The parameters settlement leads to more flexibility and reliability of the algorithm. So, settlement of the parameters is one of the important factors in reaching the optimized solution in continuous optimization problems. The population selection is very important in Meta-Heuristic algorithms. If the population number is low, the problem will soon be convergent and we will not get the favored answer or close to the global optimized answer, and if the population number is high a long time is needed for the algorithm to be convergent. So, the number of the population must be suitable and in conformity to the optimization problem to get the optimized solution. For ABC and FA to effectively search the functions space, the number of primary population is set 50 and the number of repetitions is set 100 for both algorithms. The results of Table (1) show that using the ABC and FA algorithm makes getting the optimized solution possible. So, ABC and FA are well able to find the optimized points in continuous optimization problems.

Table 1: Finding the Optimized Solution

| Function | Range of search Space | ABC | FA |
|---|---|---|---|
| $f_1$ | ±30 | 0.0059 | 0.1287 |
| $f_2$ | ±30 | 0.0136 | 0.7516 |

To show the efficiency of ABC and FA are, the convergence diagram is used. The operation of the algorithms in convergence toward the optimized answer or the suitable number of the repetitions is showed in figure (5). Studying the diagrams show that first the starting answers of the algorithms are randomly selected from the answer space and then by repeating the algorithms, the value of the goal function will get close to the optimized answer.

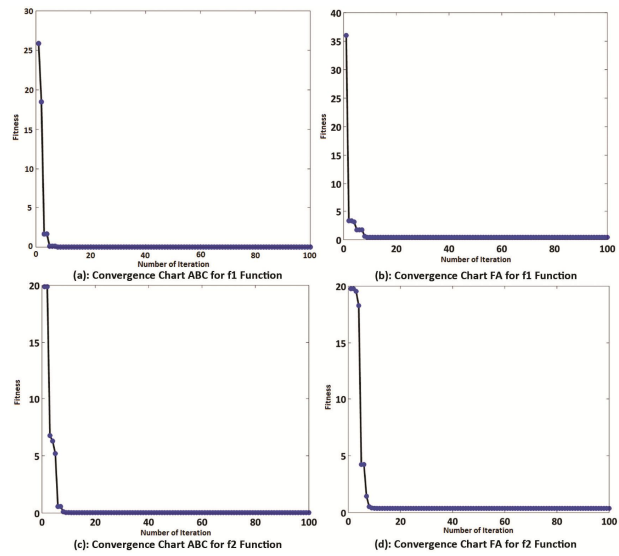

Figure 5: The Convergence Diagram of ABC and FA for Solving the Two and Three Dimensional Rastrigin Function

Table (2) shows the number of the repetition of the execution of ABC and FA are for two dimensional Rastrigin function. The reason of using the repetitive repetitions is to show the existence of probability in the structure of these algorithms. The results of Table (2) show the fact that ABC algorithm is more efficient in finding the global optimized points.

Table 2: Comparison of the Results of ABC and FA for $f_1$ function

| Algorithm | Iteration | | | | |
|-----------|------|------|------|------|------|
| | 20 | 40 | 60 | 80 | 100 |
| ABC | 1.0372 | 1.1610 | 1.0056 | 0.0148 | 0.0059 |
| FA | 1.5273 | 1.3604 | 1.1312 | 1.0228 | 0.1287 |

Figure (6) shows comparison diagram of the results of ABC and FA for $f_1$ function.

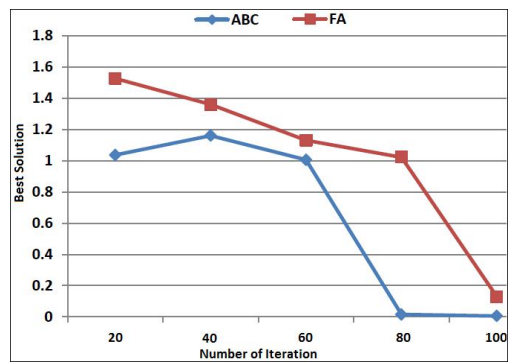

Figure 6: The Comparison Diagram of the Results of ABC and FA for $f_1$ Function

Table (3) shows the repetition number of execution of ABC and FA for three dimensional Rastrigin function. As it is seen, ABC algorithm is more efficient in solving the continuous optimization functions of large dimensions and is more able in finding the global optimized points. So, ABC algorithm is able to retain the balance of the local and global search of the problem despite the increase in the dimensions of it in an optimized way.

Table 3: Comparison of the Results of ABC and FA for $f_2$ Function

| Algorithm | Iteration | | | | |
|---|---|---|---|---|---|
| | 20 | 40 | 60 | 80 | 100 |
| ABC | 1.6723 | 1.1688 | 1.6165 | 0.0826 | 0.0136 |
| FA | 2.2512 | 2.1298 | 1.6228 | 1.1743 | 0.7516 |

Figure (7) shows comparison diagram of the results of ABC and FA for $f_2$ function.

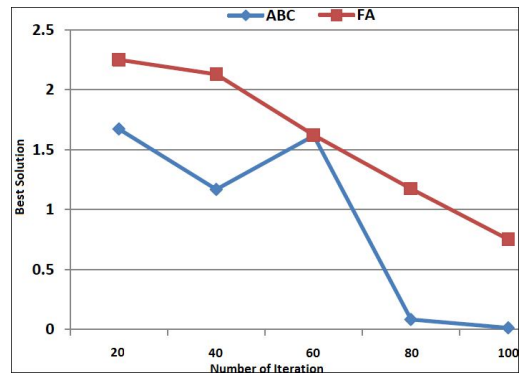

Figure 7: The Comparison Diagram of the Results of ABC and FA for $f_2$ Function

## 6. DISCUSSION

ABC and FA are efficient algorithms in different solving the continuous optimization problems. These algorithms use the quality parameters number the value of which could be settled easily. Also the speed of convergence of ABC and FA is very high in probability of finding the global optimized answer. SO, it is possible to find the optimized solution in continuous optimization problems using ABC and FA are. One of the merits of ABC algorithm is the abandonment phenomenon. It means the time the employed bees would not find the optimized solutions after some repetitions, and then they transform to the scout bees again and move in random paths to start searching for optimized solutions. By this way the solutions which are not optimized will be abandoned and again the global optimized points are searched. So, the behaviour of the bees for finding the optimized points is a combination of the two methods of local and global searching.

In comparison to the other Meta-Heuristic algorithms, ABC algorithm is a sample one to some extent because this algorithm just uses the three basic specifications of colony population, the maximum number of the repetitions and the abandonment factor. So, the implementation of ABC is very simple from calculation point of view and if the right values are used for its parameters, there would be high probability of finding the global optimized answer. For escaping the local optimization, ABC algorithm acts in a way that when it faces such location, the bees will be transferred to other parts of the search space and then will search the optimized answers there and will repeat this till reaching the global optimized answer.

ABC algorithm is well efficient in multi variable functions and also the functions which have local minimums and maximums. The movement of the employed bees to go further than the unsuitable places makes the algorithm work well facing the problems of very high dimensions and also the problems in which the population is primarily unsuitably distributed. So, the success rate of ABC algorithm in solving the optimization problems is high. The employed bees try to get close to the optimized areas an any stage in ABC and finally they try to find the optimized solutions by the help of scout bees. One of the merits of FA in solving the continuous optimization of complex functions is that it is able to change status from one optimization point to the other one. In FA, if the best solution is not found, the search is not stopped and it is done around the neighbour of the previous points to find the optimized solution. ABC and FA use the random variables in continuous optimization process. In accurate words, the answers of these algorithms are probable in their nature. In fact in these algorithms the search process takes place around the answers of the previous stage. So, these algorithms are able to escape the local optimized points. So, these algorithms must be executed repetitively to reach the optimized solution. In ABC algorithm the befitting function is used to get the optimized points and the problem space is studied more accurately by the cooperation. But in FA as the fireflies gather the more lighted firefly, it is not possible to get the optimized points well and this leads to increase of the speed of convergence to the local optimization points.

# 7. CONCLUSION AND FUTURE WORKS

In this paper we have studied the ABC and FA for solving the continuous optimization problems of vast area and the answers close to the optimized one. To evaluate the efficiency of the algorithms, two types of comparison from answer accuracy and reliability in convergence points have been studied. After studying the results, it became clear that the probability of convergence and getting the optimized answer via ABC algorithm is a little more than FA in solving the continuous optimization problems. ABC algorithm is more compatible in solving such problems and is also more powerful from answering point and when it is convergent, it is faster than FA.This is not generalizable to the other problems and such verdict is not right for all optimization problems. So, ABC algorithm is more reliable in reaching the global optimized points in comparison to FA algorithm in solving the problems like functions of continuous optimization. We hope in future will use other Meta-Heuristic algorithms for studying the continuous optimization problems of high dimensions and more optimized points presenting this paper.

## Authors

**Seyyed Reza Khaze** is a Lecturer and Member of the Research Committee of the Department of Computer Engineering, Dehdasht Branch, Islamic Azad University, Iran. He is a Member of Editorial Board and Review Board in Several International Journals and National Conferences. His interested research areas are in the Software Cost Estimation, Machine learning, Data Mining, Optimization and Artificial Intelligence.

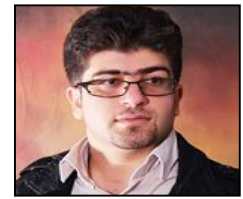

**Isa Maleki** is a Lecturer and Member of The Research Committee of The Department of Computer Engineering, Dehdasht Branch, Islamic Azad University, Iran. He Also Has Research Collaboration with Dehdasht Universities Research Association Ngo. He is a Member of Review Board in Several National Conferences. His Interested Research Areas Are in the Software Cost Estimation, Machine learning, Data Mining, Optimization and Artificial Intelligence.

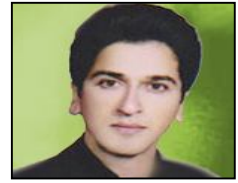

**Sohrab Hojjatkhah** is Currently Head of the Department of Computer Engineering, Dehdasht Branch, Islamic Azad University, Iran. He Has a Bachelor's Degree In Software Engineering, Received From Amir Kabir University, Iran, Then Received A Master's Degree In Artificial Intelligence From The University Of Shiraz And Currently PhD Candidate In Department Of Computer Engineering At Science And Research Branch, Islamic Azad University, Iran. His Interested Research Areas Are In The Image Processing, Speech Processing, Machine Learning, Data Mining And Artificial Intelligence.

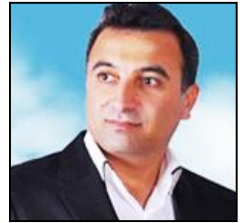

**Ali Bagherinia** is a lecturer and member of the Research Committee of the Department of Computer  Engineering, Dehdasht Branch, Islamic Azad University, Iran. He Has a Currently PhD Candidate In Department Of Computer Engineering At Science And Research Branch, Islamic Azad University, Iran. His Interested Research Areas Are in the Wireless Sensor Networks, data mining, optimization and artificial intelligence.

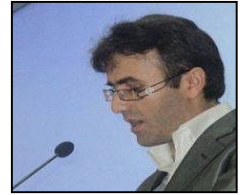